\documentclass[11pt]{article} % For LaTeX2e
\usepackage{rldmsubmit,palatino}
\usepackage{graphicx}
\usepackage[round]{natbib}  % Flexible bibliography support
\usepackage{color}
\usepackage{url}
\usepackage{graphicx}
\usepackage{amssymb}
\usepackage[utf8]{inputenc} % allow utf-8 input
\usepackage[T1]{fontenc}    % use 8-bit T1 fonts
\usepackage{url}            % simple URL typesetting
\usepackage{booktabs}       % professional-quality tables
\usepackage{amsfonts}       % blackboard math symbols
\usepackage{nicefrac}       % compact symbols for 1/2, etc.
\usepackage{microtype}      % microtypography
\usepackage{amsmath}  		% AMS mathematical facilities for LaTeX
\usepackage{amsfonts}		
\usepackage{algorithm}		
\usepackage{algorithmic}	
\usepackage{wrapfig}		
\usepackage{graphicx}		
\usepackage{subcaption}		
\usepackage{float} 			% Improved interface for floating objects
\usepackage{graphicx}		
\usepackage{caption}		
\usepackage{dirtytalk}		
\usepackage[threshold=0]{csquotes}
\setcitestyle{aysep{ }}		
\usepackage{xcolor}			
\usepackage[margin=1in]{geometry}
\usepackage[hidelinks]{hyperref}
\usepackage{dsfont}
\usepackage[bottom]{footmisc}
\usepackage{amssymb}

\DeclareMathOperator{\E}{\mathbb{E}}

\title{When is a Prediction Knowledge?}

\author{
Alex Kearney \\
Department of Computing Science\\
University of Alberta\\
Edmonton, Alberta, Canada \\
\texttt{kearney@ualberta.ca}\thanks{For additional discussion on the epistemology of predictive knowledge and how we can view predictive knowledge architectures as having semantics, please refer to Kearney \& Oxton, "Making Meaning: Semiotics Within Predictive Knowledge Architectures", RLDM 2019.
} \\
\And
Patrick M. Pilarski \\
Departments of Medicine and Computing Science\\
University of Alberta\\
Edmonton, Alberta, Canada \\
\texttt{pilarski@ualberta.ca}
}

\begin{document}

\maketitle
% Within reinforcement learning, there are the seeds of an approach to constructing machine knowledge through prediction. While promising, there is limited discussion of what the formal commitments of such an approach would be: namely, what knowledge is and what counts as it. In this paper, we take a first step towards formalizing predictive knowledge by clarifying the relationship of GVFs to formal theories of knowledge. We identify that a GVF's estimates of some cumulant can be seen as truthful insofar as they match the observed expected discounted return of the cumulant; we discuss arguments for and against the reliability of a belief---or accuracy of a prediction---as being sufficient for justifying knowledge. Having formalized these relationships between GVFs and both justification and truth, we use a robotic prediction task to demonstrate that prediction accuracy is insufficient to determining whether a prediction is knowledge. This inquiry is not simply an academic discussion: it has practical implications for decisions about what knowledge is and what counts as it in architectural proposals. The project of  predictive knowledge shows promise not just as a collection of practical engineering proposals, but also as a theory of machine knowledge; however, to achieve its full potential, predictive knowledge research must pay greater attention to the epistemic commitments being made.  
\begin{abstract}
Within Reinforcement Learning, there is a growing collection of research which aims to express all of an agent's knowledge of the world through predictions about sensation, behaviour, and time. This work can be seen not only as a collection of architectural proposals, but also as the beginnings of a theory of machine knowledge in reinforcement learning. Recent work has expanded what can be expressed using predictions, and developed applications which use predictions to inform decision-making on a variety of synthetic and real-world problems. While promising, we here suggest that the notion of predictions as knowledge in reinforcement learning is as yet underdeveloped: some work explicitly refers to predictions as knowledge, what the requirements are for considering a prediction to be knowledge have yet to be well explored. This specification of the necessary and sufficient conditions of knowledge is important; even if claims about the nature of knowledge are left implicit in technical proposals, the underlying assumptions of such claims have consequences for the systems we design. These consequences manifest in both the way we choose to structure predictive knowledge architectures, and how we evaluate them. In this paper, we take a first step to formalizing predictive knowledge by discussing the relationship of predictive knowledge learning methods to existing theories of knowledge in epistemology. Specifically, we explore the relationships between Generalized Value Functions and epistemic notions of Justification and Truth.
% Punch final sentence communicating why this is meaningful in any appreciable way. 
\end{abstract}

% Maybe it's worth emphasizing two compoenents:
%       1. it's a philosophical choice...
%       2. it's an engineering choice; we need to sift through them all...


\acknowledgements{Thanks to David Quail for insightful discussion; thanks to Dylan Jones, Kory Mathewson, and Johannes G\"unther for feedback on an early draft of this manuscript. This work was supported in part by the Canada Research Chairs program, the Alberta Machine Intelligence Institute, Alberta Innovates, the Natural Sciences and Engineering Research Council, and by Borealis AI through their Global Fellowship Award.
}  

\startmain 

\section{Predictive Approaches to Machine Knowledge}

% what is the motivation: we want to make knowledge
% one of the popular ways of doing things in reinforcement learning is predictions
% this actually isn't an unreasonable thing

One of the foundational goals of machine intelligence is to create systems which are able to understand and reason about the world around them. Within Reinforcement Learning, there is a growing collection of research which attempts to describe the world in terms of predictions about the environment, sometimes called \emph{Predictive Knowledge} \citep{sutton_grand_2009, koop_investigating_2008, sutton_horde_2011, white_developing_2015}. Predictive knowledge agents describe the world by making many predictions with respect to their behaviour. These predictions can then be interrelated to express more abstract, conceptual aspects of the environment \citep{schapire_diversity-based_1988}. For instance, using a General Value Function, a system could predict whether there is an obstacle to the left or right. Key to this approach is that all predictions---from immediate sensorimotor anticipation, to abstract conceptual expressions of the environment---are described exclusively in terms of sensation, behaviour, and time. 
% In focusing on these three constraints, predictive knowledge is liberated from one of the greatest XXXXXXX: the need for human input in developing categories in order to learn abstraction abstractions. 
As a result of these constraints, predictive knowledge centres itself around methods which are able to construct their own categories, properties and relationships: predictive knowledge is liberated from the process of labelling. This body of work can be seen as not just a collection of engineering proposals, but also as a fledgling approach to describing knowledge from a machine intelligence perspective---as a starting point for applying Epistemology to Reinforcement Learning. 

% Predictive Knowledge uses learning methods which are \emph{grounded}---methods which are able een made for the importance of prediction in how we perceive our environment \citep{gilbert_stumbling_2009}, and how we come to be aware of our environment 
% \citep{cleeremans_consciousness_2007}. Arguments drawing from neuroscience and cognitive science have impressed the importance of predicting the consequences of our behaviours in constructing perception \citep{noe_action_2004}.
% TODO: amp this up; why predictive knowledge over all the options? You can find an argument for this in your other paper...

Predictive knowledge methods show promise; however it is unclear to what extent predictions can be considered knowledge. While prediction's special status as knowledge has been alluded to in RL \citep{sutton_horde_2011,white_developing_2015}, there has been no discussion of the necessary and sufficient conditions for predictions to be considered knowledge, or the assumptions required and consequences which follow from considering predictions to be knowledge. 
This is more than simply an absence of conceptual discussion in a purely technical endeavour; there are practical challenges to developing predictive knowledge architectures which are particularly pernicious due to a limited understanding of the requirements of knowledge---i.e., how to choose \emph{what} to predict and \emph{how} to predict it independent of designer intervention is largely unknown. Although predictions have proven to be practically useful in reactive control systems in bionic limbs \citep{edwards_machine_2016} and industrial laser welding \citep{gunther_intelligent_2016}, in each of these instances the predictions learnt by the system and how they are used to inform decison-making is hand-specified by engineers and designers. These problems, at least in part, are a consequence of a poor understanding of the requirements of knowledge. 

When we propose that predictions can be interpreted as knowledge, we are making a claim about what knowledge \emph{is}. In this paper, we begin the project of formalizing a theory of knowledge in reinforcement learning by exploring justification and truth in predictive knowledge. Specifically, we 1) highlight evaluation concerns in predictive knowledge architectures, emphasizing how they relate to existing real-world applications; and 2) argue that epistemology is relevant to predictive knowledge research---that epistemology deserves greater attention when designing predictive knowledge architectures. To do so, we examine one of the most fundamental components of predictive knowledge proposals: General Value Functions (GVFs). 
% We examine whether predictions can be considered knowledge by discussing whether. 

\section{General Value Functions}

When we discuss the requirements of knowledge, it is natural for us to begin by examining how predictive knowledge learning methods relate to formal theories of knowledge. One of the central methods of specifying predictions in predictive knowledge is through General Value Functions. General Value functions estimate the discounted sum of some signal $c$ over discrete time-steps $t = 1,2,3,...,n$ defined as $G_t = \E( \sum^\infty_{k=0}(\prod^{k}_{j=1}(\gamma_{t+j}))C_{t+k+1})$. On each time-step the agent receives some vector $o_t$ of observations which describes the environment and takes an action $a_t$. The observations are used to construct the \emph{agent-state} $\phi: o_t \rightarrow \mathbb{R}^n$: the state of the environment from the agent's perspective. A GVF is parameterized by a set of weights $w \in \mathbb{R}^n$ which when combined with the agent-state produce an estimate of the return $v(s) = w^\top \phi_(o_t)$ The prediction is specified by two sets of parameters: \emph{question parameters} which determine what the prediction is about and \emph{answer parameters} which determine how the prediction is learnt. Question parameters include the signal of interest $C$, a discounting function dependent on the state of the environment $s_t$ and an action taken $a_t$, a factor $0 \geq \gamma \geq 1$ which determines how to discount future signals, and a policy $\pi$ which describes the behaviour over which the predictions are made. Answer parameters include the step-size $\alpha$ which scales updates to the weights, and the eligibility decay $\lambda$ which determines how much previous states should update their estimates based on the most recent observation. These predictions can be learned online, incrementally using policy evaluation methods such as Temporal-difference learning \citep{sutton_learning_1988-2}.

% When performing TD learning, the experience of a learning agent is described in terms of discrete time-steps $t = 0,1,2,3,...,n$. On each time-step $t$ the agent recieves a vector of observations $o_t$ and can take an action $a_t$. These observations are used to construct the \emph{agent-state} $\phi : o_t \longrightarrow \mathbb{R}^{n}$ which describes the state of the environment from the agent's perspective. The 

% These predictions \citep{sutton_horde_2011} \citep{modayil_multi-timescale_2014}

GVFs form a key component of predictive knowledge proposals by acting as the mechanism through which knowledge is constructed\citep{sutton_horde_2011}. Certainly, not all predictions are created equally. Feature construction and amount of experience contribute to the how well the return $G_t$ is estimated. If we center all knowledge as a collection of predictions, how do we evaluate the quality of a predictions as knowledge?

\section{Lessons From Epistemology: Barn Facades and Bionic Limbs}

Before embarking on determining whether or not accurate predictions can be considered knowledge, it's prudent to have an understanding of what knowledge is. To this end, we introduce arguments from epistemology, the study of knowledge, and ground these arguments in terms of GVFs. 

At its core epistemology captures the distinction between systems which \emph{know} that such-and-such is the case and systems which are simply reliably responding to stimuli. While there are many theories that define the necessary and sufficient conditions for knowledge, they can be summarized broadly as requiring Justification, Truth, and Belief  \citep{gettier_is_1963}. Each of the legs of this tripartite approach to analysing knowledge are meant to constrain what can be admitted as knowledge.

% Todo: it may be worth noting that a theory which admits everything explains nothing.
First, one must believe that they have knowledge of something. 
% This might appear to be a trivial requirement, however, there are real-world examples individuals are able to respond to stimuli.
% describes a condition where people who are blind due to brain lesions in their visual cortex are able to respond to visual stimuli which they do not perceive. 
Belief may seem trivial; however, there are real-world examples of people who are able to complete tasks while not \emph{believing} they are capable of doing so. When blindsighted patients are asked to perform certain visual tasks, they are able to achieve accuracy higher than would be expected by chance, but do not believe their reports are accurate \citep{humphrey_seeing_2006}. 
% Typically, these subjects are forced to guess; even when told of their success, blindsighted individuals do not spontaneously continue to guess without prompting .
A blindsighted person does not assert that they \emph{know} whether or not a stimulus is present; regardless, they are able to complete these tasks with some reliability. Second, the belief must be truthful. Truth separates beliefs which have bearing on the world, and assertions which are incongruous for reality. If someone says they know the moon is made of cheese, we wouldn't say they \emph{know} what the moon is made of, even if they deeply hold this belief. Third, a belief must be justified. Justification serves to separate accidentally true beliefs from those which are right for good reasons; i.e, if you asked someone how to get to the nearest cafe, and their directions happened to be correct, you wouldn't say they were right---you'd say they were \emph{lucky}.

The variety of positions relating to each of justification, truth, and belief are numerous. To that end, we constrain ourselves to considering how GVFs relate to the first two components of the tripod: how can predictions be licensed as being Truthful and Justified? As we alluded to earlier, not all predictions are created equally. In order to make progress in designing predictive architectures, we must be able to separate predictions which are unreliable, or made for poor reasons, from those which are robust and can be used to inform decision-making.
% One might be able to assert that a prediction is, in a sense, a belief. 

When an agent is making a prediction, it is making an assertion about the world as observed through its data stream. If a prediction is accurate, it is a testament to its truth. One common method of evaluating whether a prediction is correct or not is to compare what is predicted against an estimation of the true return \citep{pilarski_dynamic_2012,edwards_machine_2016,gunther_intelligent_2016}. The approximate return is $ \tilde{G_t} = \sum^\textit{b}_{k=0}(\prod^{k}_{j=1}(\gamma_{t+j}))C_{t+k+1} - v_{t}(s_{t})$ for some buffer-size $b$ which determines how many steps into the future cumulants $c$ are stored to produce the return estimate on any given time-step. The truthfulness of the prediction can be described as the the extent to which estimated value matches the true, observed return\footnote{This approach is advocated in the original proposal of \cite{sutton_horde_2011}}.

% Such on-line evaluation methods can be extended to off-policy learning settings by taking \emph{excursions}---by treating the target policy of a prediction as the behaviour policy in order to collect the experience required to calculate the estimated return for a set of predictions \citep{excursions}.

If prediction accuracy describes the truth of a prediction, what is justification within predictive knowledge architectures? Or, is justification necessary? As previously mentioned, the necessary and sufficient conditions for knowledge are a point of contention. In the same paper that \citeauthor{gettier_is_1963} introduced Justified True Belief, he argued against its validity. Similarly, \citeauthor{goldman_discrimination_1976}'s Barn Facade problem---which we explore in terms of predictions in the following paragraphs---illustrates how evidence and reasons are not the only way to support the claim that a belief is true---reasons are not the only way to separate a lucky guess, from a justified belief \citep{goldman_discrimination_1976}. The purpose of justification is simply to show that a belief is expected to be reliable, that a belief is predicted to be true \citep{brandom_articulating_2009}. Can we treat the reliability of predictions as sufficient for identifying knowledge independent of any other form of justification?

In short, no. While the reliability of a belief---or, accuracy of a prediction---is a means of justifying a belief, reliability alone is insufficient to attribute knowledge \citep{brandom_articulating_2009}. We can examine the limitations of reliability as justification by translating \citeauthor{goldman_discrimination_1976}'s barn facade problem to a predictive knowledge experiment. Consider a single GVF making a prediction about some signal $c$. In this case, the return error $v_t(\phi_t) - \tilde{G_t}$ is relative to a particular time-step $t$, and a set of observations $o_t$. A GVF which predicts random values could make a perfect prediction for a given time-step $t$ and have no return error for an observation $o_t$. Clearly, the accuracy of a prediction over one time step says nothing about how likely a prediction is to be accurate in general. Given the limitations of a single time-step error, over what horizon---for what period of time, or what collection of states---must we examine the return error to licence truth?
% That is, it says nothing of the global error. 
Must we calculate the return error of a prediction relative to all possible states in order to determine whether a prediction is sufficiently justified? Such a requirement would be technically infeasible in a real-world setting. 

Not only is return error impractical as the exclusive source of justification, it is incoherent on a conceptual level. Relative to each set of states, there is a clear answer as to whether or not a prediction is accurate; however, there is nothing in the world which privileges one set of states over others in making the distinction of truth. So the accuracy, or reliability of a belief does not determine whether or not the prediction is justified. None of this is to say that the reliability of predictions does not have any epistemic significance. Prediction accuracy is unquestionably an important part of assessing the truth of a prediction and evaluating if a prediction is justified. However, Prediction accuracy alone does not tell the full story.

\begin{figure*}[t!]
    \centering
    \begin{subfigure}[b]{0.5\textwidth}
        \centering
        \includegraphics[ width=\linewidth,keepaspectratio]{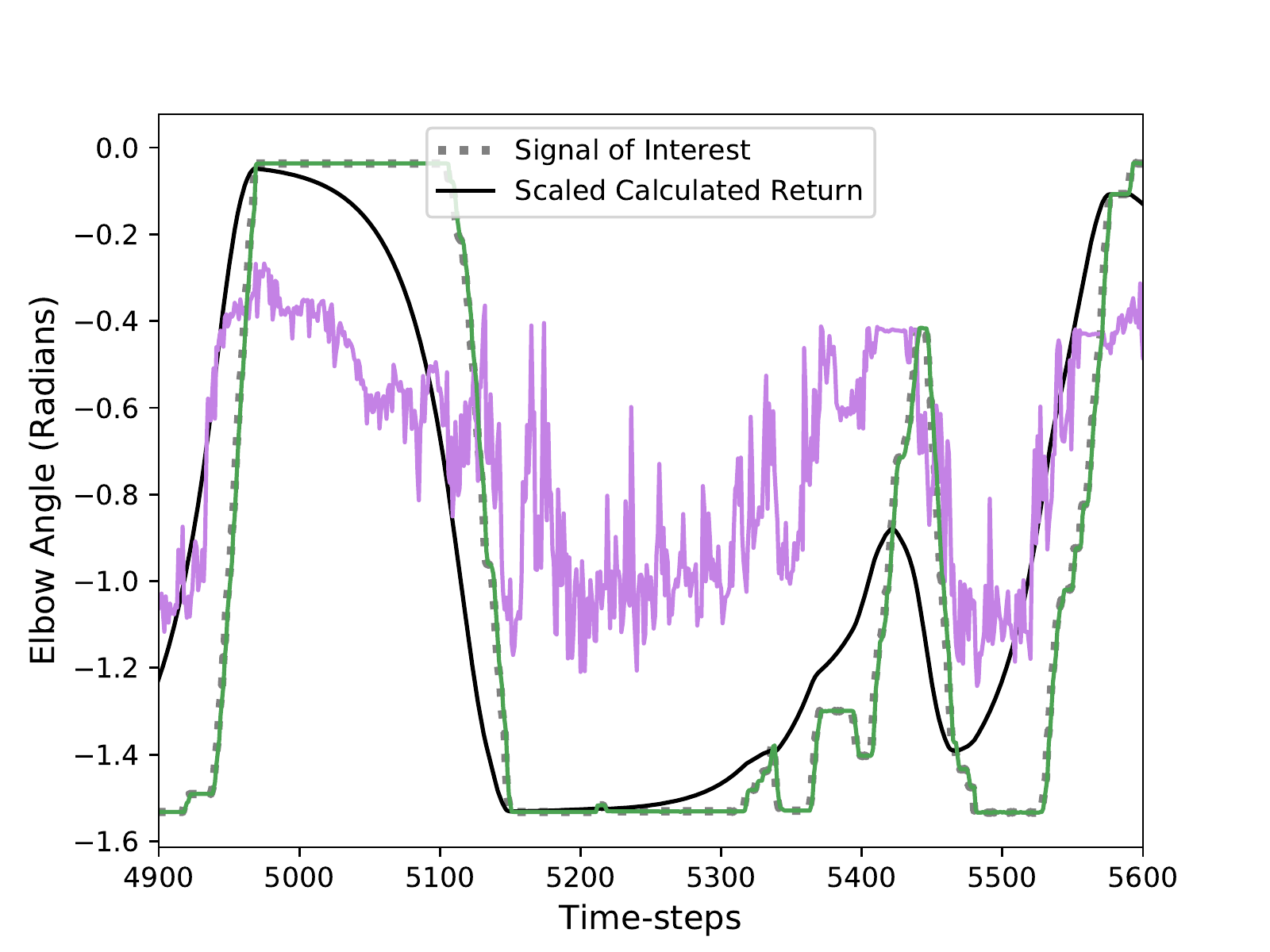}
        \caption{Estimated value for two predictions; Signal as dotted line.}
        \label{prediction}
    \end{subfigure}%
    ~ 
    \begin{subfigure}[b]{0.5\textwidth}
        \centering
        \includegraphics[ width=\linewidth,keepaspectratio]{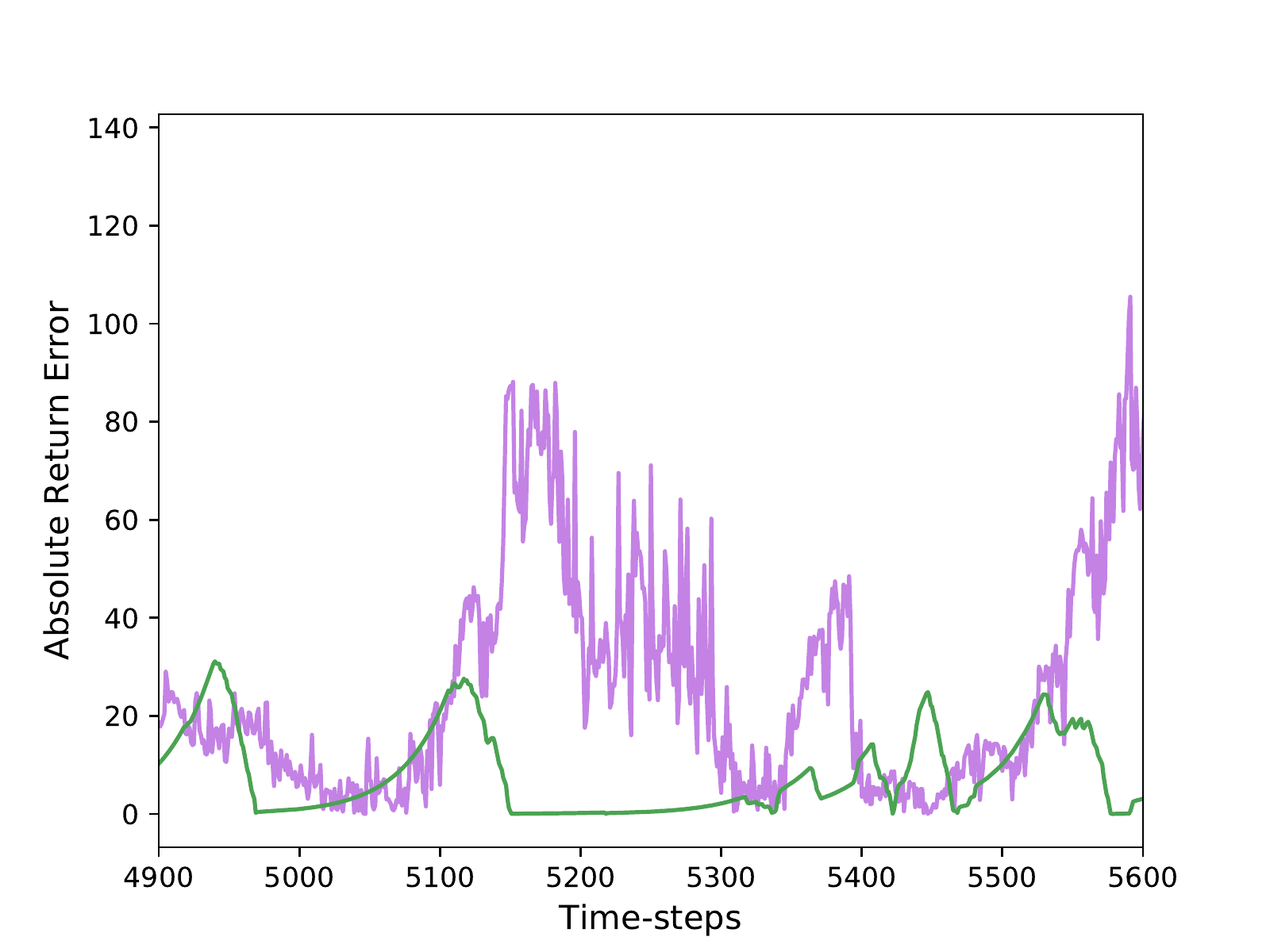}
        \caption{Absolute return error of two predictions.}
        \label{error}
    \end{subfigure}
    \caption{Which prediction counts as knowledge: green or purple?}
    \label{comparison}
    \vspace{-2em}
\end{figure*}

% as sufficient for describing knowledge. In short, no. If one has no notion of justification, one cannot have an understanding of reliability, nor knowledge \cite{brandom_articulating_2009}. If a system is simply able to reliably indicate and discriminate, then one cannot say it knows. A chunk of iron is able to classify environments with and without oxygen by rusting, but rusting is not a conceptual or cognitive process; rusting is simply a reliable response to environmental stimulus.

To explore this point, we produce two predictions about the joint angles of a robotic actuator, as sampled from the human  controlling a robotic arm to do manipulation task. Please refer to \citet{pilarski_adaptive_2013} for the full details of how this dataset was generated. %The cumulant is the encoder position of the shoulder servo on the BentoArm \citep{dawson_development_2014}: a bionic limb. 
The cumulant of interest is the elbow servo motor angle in radians. For both predictions, the discount factor is $ \gamma = 0.99$, corresponding to roughly 2.5 seconds of arm operation. As per \citet{pilarski_adaptive_2013}, the predictions were made on-policy with TD($\lambda$) with $\lambda=0.999$ and a step size $\alpha= 0.033$ \citep{sutton_learning_1988-2}. Only the function approximators used to construct the agent-state varies between the two predictions. 

From the predictions in Figure \ref{comparison}, we can see that the green prediction isn't a prediction at all. Although both predictions are specified to learn the same GVF, the green prediction is simply tracking the signal of interest. In comparison, the purple prediction, in fact, predicts: it rises before the stimulus rises, and decreases before the stimulus falls. Looking at return error alone (Figure \ref{error}), we would be lead to the conclusion that the green prediction is in fact more truthful than the purple. Because the green prediction is more accurate---both on a moment-to moment basis, and throughout the trial---from this \emph{reliabilist} perspective, it is better justified. We could conclude that the green prediction that isn't predicting is a better candidate for knowledge. Although the purple prediction is clearly more predictive, it has a greater return error, both on a moment-to-moment basis on each time-step and in the greater context of the experimental trial.

More than simply a contrived example, these predictions are examples of prototypical GVFs made on bionic limbs to inform control systems. While existing systems are hand-engineered, if we choose to build systems which independently make decisions about what to learn and how to learn them, we must be able to assess the quality of a prediction in a robust, reliable way. From purely an engineering standpoint, in order to build such systems successfully we must be able to discriminate between predictions which have low error for poor reasons and predictions which explain their signal of interest \citep{pilarski_real-time_2013}. Put simply, just because a prediction is accurate, doesn't make it useful. 

% Regardless of whether or not we acknowledge epistemology in the design decisions when we attempt to build systems which have knowledge.
The limitations of reliability as justification is more than a conceptual problem, it has practical consequences for evaluation in real-world applications of predictive knowledge systems. The consequences of epistemic choices we make---whether we are conscious of them or not---have a fundamental impact on the effectiveness of our systems. To achieve its fullest potential, future work should examine additional methods of supporting the justification of predictions, perhaps using internal signals about learning.

\section{Concluding Thoughts:\\The Importance of Evaluating When Predictions are Knowledge}
Within reinforcement learning, there are the seeds of an approach to constructing machine knowledge through prediction. While promising, there is limited discussion of what the formal commitments of such an approach would be: namely, what knowledge is and what counts as it. In this paper, we take a first step towards formalizing predictive knowledge by clarifying the relationship of GVFs to formal theories of knowledge. We identify that a GVF's estimates of some cumulant can be seen as truthful insofar as they match the observed expected discounted return of the cumulant; we discuss arguments for and against the reliability of a belief---or accuracy of a prediction---as being sufficient for justifying knowledge. Having formalized these relationships between GVFs and both justification and truth, we use a robotic prediction task to demonstrate that prediction accuracy is insufficient to determining whether a prediction is knowledge. This inquiry is not simply an academic discussion: it has practical implications for decisions about what knowledge is and what counts as it in architectural proposals. The project of  predictive knowledge shows promise not just as a collection of practical engineering proposals, but also as a theory of machine knowledge; however, to achieve its full potential, predictive knowledge research must pay greater attention to the epistemic commitments being made.  

\vspace{-1em}

\footnotesize
\bibliography{My_Library.bib}
\end{document}